# A Common-Factor Approach for Multivariate Data Cleaning with an Application to Mars Phoenix Mission Data

Dongping Fang
Predictive Analytics Center of Excellence
Zurich North America
dongping.fang@zurichna.com

Elizabeth Oberlin
Department of Chemistry
Tufts University
Elizabeth.Oberlin@tufts.edu

Wei Ding
Department of Computer Science
University of Massachusetts Boston
wei.ding@umb.edu

Samuel P. Kounaves
Department of Chemistry
Tufts University
Samuel.Kounave@tufts.edu

**ABSTRACT**

Data quality is fundamentally important to ensure the reliability of data for stakeholders to make decisions. In real world applications, such as scientific exploration of extreme environments, it is unrealistic to require raw data collected to be perfect. As data miners, when it is infeasible to physically know the why and the how in order to clean up the data, we propose to seek the intrinsic structure of the signal to identify the common factors of multivariate data. Using our new data-driven learning method—the common-factor data cleaning approach, we address an interdisciplinary challenge on multivariate data cleaning when complex external impacts appear to interfere with multiple data measurements. Existing data analyses typically process one signal measurement at a time without considering the associations among all signals. We analyze all signal measurements simultaneously to find the hidden common factors that drive all measurements to vary together, but not as a result of the true data measurements. We use common factors to reduce the variations in the data without changing the base mean level of the data to avoid altering the physical meaning.

We have reanalyzed the NASA Mars Phoenix mission data used in the leading effort by Kounaves's team (lead scientist for the wet chemistry experiment on the Phoenix) [1, 2] with our proposed method to show the resulting differences. We demonstrate that this new common-factor method successfully helps reducing systematic noises without definitive understanding of the source and without degrading the physical meaning of the signal.

## 1. INTRODUCTION

Data quality is fundamentally important for domain scientists. In real world applications, such as scientific exploration of extreme environments, it is unrealistic to require data collection to be perfect. Our motivating application is to recover the true chemical analysis data from the Wet Chemistry Laboratory (WCL) on the 2008 Phoenix Mars Lander (Fig. 1 & [1]). The WCL collected over three-million data points and performed the first comprehensive wet chemical analysis of the soil on Mars. The initial data has provided new scientific insights into the history of Mars, its potential for supporting microbial life, and even its atmospheric chemistry, with resulting publications in *Science* [4]. Six years later less than 1% of the WCL data has been manually studied [2]. The main reason for this is the noise introduced by unexpected instrumental and environmental factors on Mars make data interpretation an ill-posed problem if data cleaning solely relies on the chemical and physical models understood on Earth.

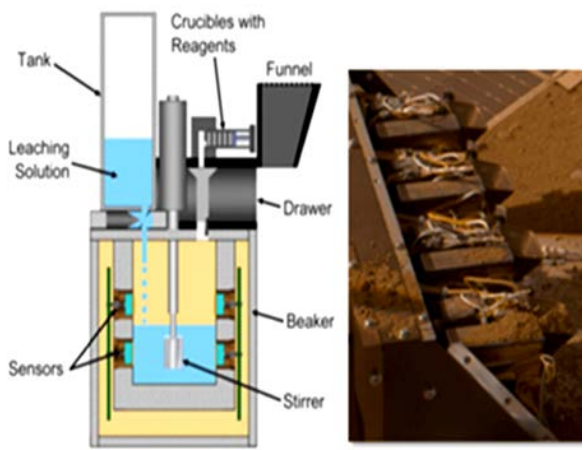

**Figure 1. Left: Schematic diagram of the WCL interior with various components (not to scale); Right: The WCL on Mars after the first analysis [2].**

As data miners, when it is infeasible to physically know the why and the how in order to clean up the data, we propose to seek the intrinsic structure of the signal to identify the common factors of multivariate data. The existing state-of-the-art denoising methods, including Fourier filtering [8], Kalman smoothing [9] [12], Gaussian smoothing [10], and Hidden Markov Model denoising [11], do not work well for this type of problem because they are designed to cope with a single data measurement, ignoring the associated shared behavior among multiple datasets. In [13-16], the principles of identifying common-mode regularities were discussed to identify a simplifying structure, shared trends in financial data and co-variance selection in biological data, but not for the data cleaning using common external factors which is a quite different problem.

We study the intrinsic characteristics of the data and propose a new common-factor removal method that utilizes multiple sensor measurements simultaneously to find the hidden shared factors which drive all measurements to vary simultaneously. These common factors represent the errors and variations caused by the combined and complicated influence of common varying external factors. We iteratively estimate the common factors by minimizing the sum of squared errors of all the sensor data. We then clean the data by removing the effects of these common factors (details in Section 2). We compare our proposed method with state-of-the-art data denoising methods (details in Section 3). We reanalyze the WCL data used in the leading effort by Kounaves et al. [2] with our proposed method to show the data quality improvement (details in Section 4).

The contribution of this paper is that we address an interdisciplinary challenge to provide a new and physically meaningful data cleaning method to improve data quality in scientific data. The proposed common-factor data cleaning approach is designed for a scenario when multiple data measurements are impacted together by an unknown and hard-to-reproduce real-world environment. In the Martian data analysis, we demonstrate that this new common-factor method can help reduce systematic noise without definitive understanding of the source and without degrading the physical meaning of the signal. Our results successfully lead to a more accurate measurement of the soil chemistry on. Though the idea of common factors is used in many fields [5, 13 – 16], to the best of our knowledge, we are the first research team to use the idea of common hidden factors on data cleaning.

## 2. METHODOLOGY

Let us assume that there are I signal sensors to collect data simultaneously. In the WCL data, these are I Ion Selective Electrode (ISE) sensors in the same beaker measuring various ions of interest. Let $E_t^{(i)}$ denote the measured value for signal $i \in \{1, ..., I\}$ at time $t \in \{t_1, ..., t_n\}$. So the observed data are $\{t, E_t^{(1)}, ..., E_t^{(I)}\}_{t=t_1}^{t_n}$. In an ideal world of data collection, each signal data measurements over time should be a constant plus a random measurement error, i.e.

$$E_t^{(i)} = \mu^{(i)} + \varepsilon_t^{(i)}, \quad (1)$$

where $\varepsilon_t^{(i)}$ denotes the measurement error for signal $i$ at time $t$, $\mu^{(i)}$ denotes the constant representing real potential for signal $i$. The $\mu^{(i)}$ does not change with time, and $\varepsilon_t^{(i)}$ is a random white noise so its value at different times or for different signals are independent, it follows

$$\text{corr}\left(\varepsilon_{t_1}^{(i_1)}, \varepsilon_{t_2}^{(i_2)}\right) = 0 \text{ for } i_1 \neq i_2 \text{ or } t_1 \neq t_2$$
$$Var\left(\varepsilon_t^{(i)}\right) = \sigma_i^2 > 0$$

(2)

If this is the case, the estimated measurement and associated error would simply be data mean $\hat{\mu}^{(i)} = \overline{E^{(i)}} = \frac{1}{n}\sum_{k=1}^{n} E_{t_k}^{(i)}$ and standard deviation of mean $\hat{\sigma}_i\sqrt{\frac{1}{n}}$ where $\hat{\sigma}_i^2 = \frac{1}{n-1}\sum_{k=1}^{n}(E_{t_k}^{(i)} - \overline{E^{(i)}})^2$.

But in reality when complex external impacts appear to interfere with multiple data measurement, we will observe deviations from the ideal case. For example, as illustrated in Fig.2 of Martian soil data, different signal measurements were correlated, exhibiting systematic fluctuations.

Our data-cleaning goal, formally speaking, is to remove the deviations to regain the forms of Eqs. (1) and (2).

**Formulation and Algorithm.** Let $K$ denotes the number of common factors, $F_{kt}$ the $k^{th}$ common factor at time $t$. The observed data can be modeled as

$$E_t^{(i)} = \mu^{(i)} + \beta_1^{(i)} F_{1t} + \cdots + \beta_K^{(i)} F_{Kt} + \varepsilon_t^{(i)}$$

(3)

where $\beta_1^{(i)}, ..., \beta_K^{(i)}$ are the coefficients of the $K$ common factors for signal $i$, and $\varepsilon_t^{(i)}$ are random noise as in Eq. (2). Notice that the common factors are the same for all multivariate data, but their influences on each signal may be different due to its different physical properties which is reflected in the coefficients $\beta^{(i)}s$ for that signal. We want to use common factors to help us reduce the variations in the data without changing the base mean level of the data. So we require the base mean of factors to be zero.

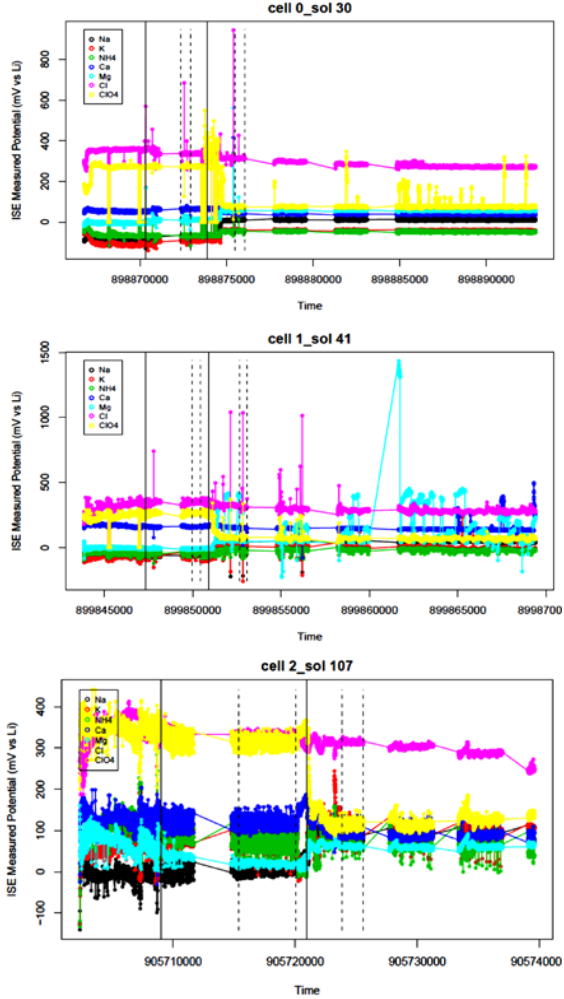

**Figure 2.** $K^+$, $Na^+$, $Mg^{2+}$, $Ca^{2+}$, $NH_4^+$, $Cl^-$, and $ClO_4^-$ sensor data (better viewed in color). The two solid vertical lines indicate the addition of the calibrant crucible and the soil sample. First pair of dashed vertical lines marks the Calibrant Interval and second pair marks the Soil Sample Interval. Data in these two intervals will be used to do further analyses. Time is spacecraft clock (SCLK) time in seconds. Top: Cell 0 data on Martian Solar Day sol 30, Middle: Cell 1 data on sol 41, Bottom: Cell 2 data on sol 107.

The cleaned data to be calculated are

$$E_t^{*(i)} = E_t^{(i)} - \beta_1^{(i)} F_{1t} - \cdots - \beta_K^{(i)} F_{Kt} = \mu^{(i)} + \varepsilon_t^{(i)} \quad (4)$$

**Parameter Estimation.** Let observed data matrix of ISE sensor measurements be

$$\mathbb{E}_{n \times I} = \begin{pmatrix} E_{t_1}^{(1)} & \cdots & E_{t_1}^{(I)} \\ \vdots & \ddots & \vdots \\ E_{t_n}^{(1)} & \cdots & E_{t_n}^{(I)} \end{pmatrix} = \begin{pmatrix} \boldsymbol{E'}_{t_1} \\ \vdots \\ \boldsymbol{E'}_{t_n} \end{pmatrix} = \left(\boldsymbol{E}^{(1)}, \ldots, \boldsymbol{E}^{(I)}\right),$$

where $\boldsymbol{E'}_{t_l}$, $\boldsymbol{E}^{(i)}$ are the $l^{th}$ row and $i^{th}$ column vectors of $\mathbb{E}$.

The model parameters are

$$\mathbb{B}_{I \times K} = \begin{pmatrix} \beta_1^{(1)} & \cdots & \beta_K^{(1)} \\ \vdots & \ddots & \vdots \\ \beta_1^{(I)} & \cdots & \beta_K^{(I)} \end{pmatrix} = (\boldsymbol{\beta}_1, \ldots, \boldsymbol{\beta}_K),$$

$$\mathbb{F}_{n \times K} = \begin{pmatrix} F_{1t_1} & \cdots & F_{Kt_1} \\ \vdots & \ddots & \vdots \\ F_{1t_n} & \cdots & F_{Kt_n} \end{pmatrix} = (\boldsymbol{F}_1, \ldots, \boldsymbol{F}_K),$$

$$\boldsymbol{\mu} = (\mu^{(1)}, \ldots, \mu^{(I)})', \Sigma = \text{diag}(\sigma_1^2, \ldots, \sigma_I^2),$$

where $\boldsymbol{\beta}_k$ and $\boldsymbol{F}_k$ are the $k^{th}$ column vector of coefficient matrix $\mathbb{B}$ and factor matrix $\mathbb{F}$ respectively and $\Sigma$ is the diagonal variance matrix.

Given observed data $\mathbb{E}$, our goal here is estimate $\mathbb{B}$, $\mathbb{F}$, $\boldsymbol{\mu}$ and $\Sigma$ by minimizing sum of squared errors in Eq. (5)

$$S = \sum_{i=1}^{I} \sum_t \left(E_t^{(i)} - \hat{E}_t^{(i)}\right)^2 = \sum_t s_t = \sum_{i=1}^{I} S^{(i)} \quad (5)$$

where $\hat{E}_t^{(i)} = \mu^{(i)} + \beta_1^{(i)} F_{1t} + \cdots + \beta_K^{(i)} F_{Kt}$, and

$$s_t = \sum_{i=1}^{I} \left(E_t^{(i)} - \hat{E}_t^{(i)}\right)^2 \quad (6)$$

$$S^{(i)} = \sum_t \left(E_t^{(i)} - \hat{E}_t^{(i)}\right)^2 \quad (7)$$

Starting from an initial value of $\boldsymbol{\theta} = (\mathbb{B}, \boldsymbol{\mu}, \Sigma)$, we will do the minimization by alternately performing estimate factors $\mathbb{F}$ given model parameters $\boldsymbol{\theta}$, and estimate $\boldsymbol{\theta}$ given $\mathbb{F}$. Algorithm 1 describes the common factor approach.

---

**Algorithm 1** Common-factor Learning with Least Square Regression

**Step I. Initialization.**
Apply statistical factor analysis to get initial estimates for $\mathbb{B}$, $\boldsymbol{\mu}$, and diagonal variance matrix $\Sigma$.

**Step II. Iteration: repeat 1 and 2 until converge.**
1. Estimate $\mathbb{F}$ for a given $\boldsymbol{\theta} = (\mathbb{B}, \boldsymbol{\mu}, \Sigma)$.
   a) For each $t$, perform weighted least square regression of $(\boldsymbol{E}_t - \boldsymbol{\mu})$ on $\boldsymbol{\beta}_1, \ldots, \boldsymbol{\beta}_K$ to get new estimates of common factors. This gives

   $$\mathbb{F}' = (\mathbb{B}'\Sigma^{-1}\mathbb{B})^{-1} \mathbb{B}'\Sigma^{-1}(\mathbb{E} - \mathbb{U})',$$

   where $\mathbb{U} = \begin{pmatrix} \mu^{(1)} & \cdots & \mu^{(I)} \\ \vdots & \ddots & \vdots \\ \mu^{(1)} & \cdots & \mu^{(I)} \end{pmatrix}$.

   b) Set trimmed mean of each factor to zero. For each factor, let
   $$\boldsymbol{F}_k \leftarrow \boldsymbol{F}_k - m_k,$$
   where $m_k$ is the trimmed mean of $k^{th}$ factor calculated by
   $$m_k = \text{mean}\{F_{tk}: |F_{tk} - \text{mean}(\boldsymbol{F}_k)| \leq c \cdot \sigma(\boldsymbol{F}_k)\}$$

   Therein $\sigma(\boldsymbol{F}_k)$ is the standard deviation of $\boldsymbol{F}_k$, and $c$ is a constant.

2. Estimate $\boldsymbol{\theta}$ for given $\mathbb{F}$.
   For given $\mathbb{F}$, fit the model in Eq. (3) for each ion $i$ by least square linear regression, i.e. regress $\boldsymbol{E}^{(i)}$ on $\boldsymbol{F}_1, \ldots, \boldsymbol{F}_K$. This step gives the new estimates for $\boldsymbol{\theta} = (\mathbb{B}, \boldsymbol{\mu}, \Sigma)$.

---

Step I initializes the algorithm by the statistical factor analysis [5]. It produces reasonably good initial values for $\mathbb{B}$ before our search starts, but it doesn't minimize the sum of squared errors for the ISE signals and thus cannot fulfill our goal. We need step II to iteratively perform the minimization.

Step II.1 uses estimated coefficients as known to estimate factor scores by minimizing $s_t$ in Eq. (6) for each $t$, which results in the ordinary least square regression. Since each signal may have different variance we modify this step using weighted least square regression instead of the ordinary least square regression. Step II.2 uses estimated factors as knowns to get a new estimate of coefficients by minimizing $S^{(i)}$ in Eq. (7) for each signal *i*. We alternately use Step II.1 and Step II.2 until the sum of squared error stops decreasing.

Step II.1.b) makes sure the trimmed mean of each factor is zero to serve the goal of cleaning up the variation part but not the base mean level part of the data. Without a good reason, the data base mean level should not be altered by any data cleaning method because it would change the physical meaning of the data. Common factors are capable of finding spikes (see factor plots in Fig. 7). We set the trimmed mean, instead of the regular mean, to be zero to reduce the influence of large outliers (spikes) on the base mean level of factors and in turn on the base mean of the data. In our calculation of the trimmed mean we suggest to statistically choose c = 2.326 which corresponds to 99%-percentile of standard normal distribution.

**Determination of Number of Common Factors.** When the number of factors increases, the total sum squared error in Eq. (5) will decrease. In the extreme case, the errors would decrease to zero if the number of factors is greater than or equal to the number of variables in the observed data. Our purpose is to use common factors to clean the data by removing influences that are believed common to variations of all ISE measurements, not the random intrinsic measurement error, $\boldsymbol{\varepsilon}_t^{(i)}$ in Eq. (1), which every device independently has. The key word here is *common*. So if adding a factor only decreases error of a single ISE measurement, this is not considered a common factor and is not used in the method. Our strategy is to try a range of number of factors starting from 0 factors, and stop when no significant multiple error decreases are observed. For example, the search will stop if the decrease of $\boldsymbol{\sigma}_i^2$ is bigger than some critical value only in one ion.

## 3. COMPARATIVE STUDIES USING SYNTHETIC DATA

Our goal of multivariate data cleaning is to identify the common variation part in the multiple series and regain the original data distribution in the forms of Eqs. (1) and (2). It leaves the measurement errors (white noise) alone while providing good estimates of the true mean and standard deviation of the estimates. In order to evaluate the performance the common-factor cleaning method, we compare it with commonly used methods of data cleaning for scientific applications including the Fourier filtering approach and Kalman smoother approach. The Fourier filtering method gives the estimate of the underlying mean but no associated errors. The Kalman smoother gives both mean and associated errors but needs a model, if the model is Eq. (1), the Kalman smoother just estimates the mean and prediction error by the data mean and standard deviation of the mean.

In the existing Marian soil data analysis, Kounaves et al. [2] used Fourier filtering method to get rid of the high frequency variations in WCL data. Toner et al. [3] used Kalman smoothing method with random walk plus noise model: $E_t = \mu_t + \varepsilon_t$ & $\mu_t = \mu_{t-1} + \eta_t$. In our comparative studies, we keep the same settings for both methods.

**Data Generation and Distribution.** In our experiments, we simulate three independent series from model $\mu + \varepsilon_t$ with $\varepsilon_t$ following a standard normal distribution $\varepsilon_t \sim N(0,1)$ and $\mu = 1, 5$ & 10 respectively for the three series, each with 100 data points (Fig. 3a). We then add 1 common factor $F_t$ (Fig. 3b) to contaminate series 1 to 3 in increasing degree ($F_t$, $1.5F_t$ & $2F_t$) to simulate the observed series (Fig. 3c). We then apply on the contaminated series the Fourier filtering (Fig. 3d), Kalman smoother (Fig. 3e) and common-factor method (Fig. 3f) to clean the data. Fig. 3 clearly shows the difference of the common-factor cleaning method from the other methods, the white noises are still in after common-factor cleaning. Figs. 3a and 3f illustrate the strong similarity between the common-factor cleaned data and the original uncontaminated data, while Fourier filter and Kalman smoother (Figs. 3d and 3e) fail to recover the true data distribution.

**Mean and Associated Error.** Kalman smoother gives both mean and associated prediction error, we compare the common-factor cleaning method and the Kalman smoother method in terms of finding the mean and associated error. To do this, we perform the above mentioned simulation 1,000 times and apply the cleaning methods and then calculate the mean and associated error for each method. Hence we have 1,000 estimated means and standard errors (distribution of these 1000 estimated means and standard errors are in Figs 4 & 5). For simplicity, the associated error for Kalman smoother here is taken as the square root of the mean Kalman variance which is smaller than that used in Toner et al. [3] because they also included the variation in mean estimates at different time points.

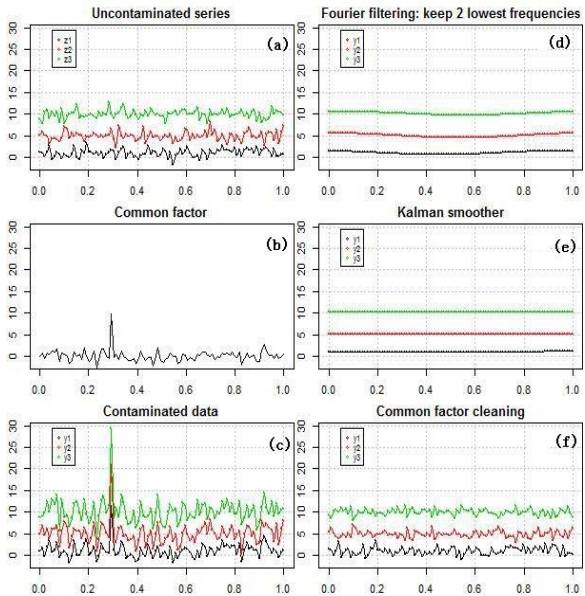

**Figure 3. (a) uncontaminated series, (b) common-factor series, (c) contaminated series, (d) Fourier filtering, (e) Kalman smoother, and (f) common-factor cleaned series**

**Estimation of Mean**. For the three uncontaminated series, their true means are 1, 5 & 10 respectively. Figure 4 shows that the Kalman smoother cleaned series (Fig. 4c) is very similar to that of the contaminated series (Fig. 4b) instead of the uncontaminated series (Fig. 4a). On the other hand, the common-factor cleaned series is more like the uncontaminated series: both centered on the same true mean (Figs. 4a & 4d). This shows the common-factor cleaning method's ability to remove systematic deviations while leaving the true mean unaffected.

**Estimation of Standard Deviation**. For the three uncontaminated series, the true standard deviations of 100 data point means are 0.1. Figure 5 shows the distribution of the standard deviation of the mean. The contaminated series (Fig. 5b) and the Kalman smoother cleaned series (Fig. 5c) are both centered on similar values which are much larger than the true value. The common-factor cleaned data (Fig. 5d) have a standard error much closer to the true value (Fig. 5a), but it underestimates the error in series with larger variations.

## 4. ANALYSIS OF MARTIAN SOIL DATA

Our evaluation using the simulated data are consistent with the assumption that the common variations between the individual data sets can be isolated and removed by analyzing the behavior of the data sets as a unit. In this section, we apply the common-factor cleaning method to the Wet Chemistry Laboratory (WCL) data from Phoenix lander mission.

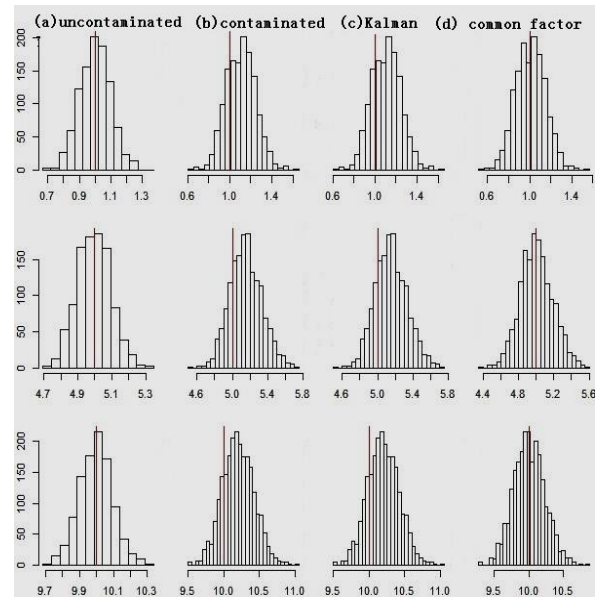

**Figure 4. Histograms of series mean. (a) uncontaminated, (b) contaminated, (c) Kalman smoother cleaned, and (d) common-factor cleaned series. Top to bottom: series 1 to 3. Red vertical line is the true mean.**

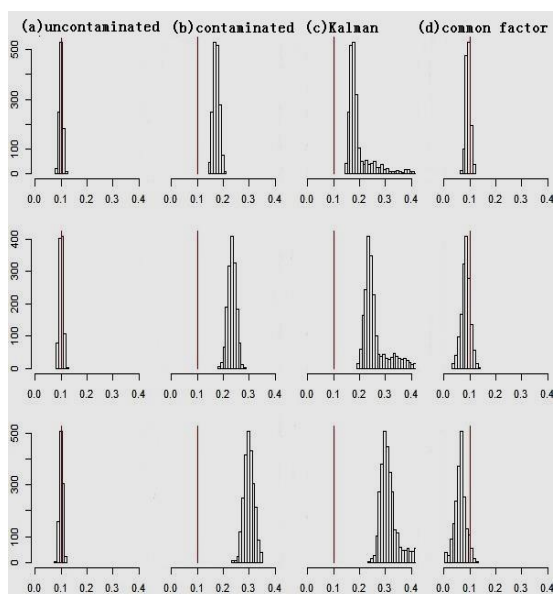

**Figure 5.** Histograms of standard deviation of the estimated mean. (a) uncontaminated, (b) contaminated, (c) Kalman smoother cleaned, and (d) common-factor cleaned series. Top to bottom: series 1 to 3. Red line marks the true standard deviation of the mean

### 4.1 WCL data from Phoenix lander

The Wet Chemistry experiments used four identical WCL cells, Cells 0, 1, 2 and 3, to analyze the soluble contents of the Martian regolith. The analyses sampled Martian regolith from four separate locations on four separate Martian solar days (sols). Cell 0 analyzed the sample "Rosy Red" taken from the surface of the Burn Alive trench; and Cells 1 and 2 analyzed samples "Sorceress 1" and "Sorceress 2", respectively, both taken from adjacent locations at a depth of ~5cm, in contact with the ice table of the Snow White trench. Sample delivery to Cell 3 failed, and the data returned was used as an in situ blank against which the remaining three analyses could be compared. Figure 1 shows a schematic diagram of the WCL cell and how the four cells looked outside after the first-day analysis on Mars. Each cell consisted of (1) an upper actuator assembly with a drawer for adding soil, "leaching solution", five crucibles with reagents used for the WCL calibration, and a stirrer; and (2) a lower beaker lined with an array of sensors including ion selective electrodes (ISE) for measuring $K^+$, $Na^+$, $Mg^{2+}$, $Ca^{2+}$, $NH_4^+$, $Ba^{2+}$ (for $SO_4^{2-}$), $Cl^-$, $Br^-$, $I^-$, $NO_3^-/ClO_4^-$, $H^+$(pH), $Li^+$; and electrodes for measuring conductivity, redox potential, cyclic voltammetry (CV), chronopotentiometry (CP), and an $IrO_2$ pH electrode. These data are publicly available at the NASA Planetary Data System [6, 7].

In this paper we only use the potential readings from the ISEs for $Na^+$, $K^+$, $Ca^{2+}$, $Mg^{2+}$, $Cl^-$, $ClO_4^-$, and $Li^+$ obtained from the analyses performed in Cell 0 on sol 30 (the 30$^{th}$ Martian solar day of the 152-sol Phoenix surface mission), Cell 1 on sol 41, and Cell 2 on sol 107. The data, prior to application of our common-factor algorithm, from these cells is displayed in Fig. 2. The time intervals chosen for analysis correspond to the originally analyzed time-series as these represent the most stable and reliable portions of the data, and provide us the ability to compare our results with the originally published values. For each Cell two regions, confined between each set of vertical dashed lines in Fig. 2, were treated with our common-factor algorithm. The first region represents the calibration interval during which the ISEs were calibrated using a solution of known concentration, described in more detail elsewhere [2]. The second interval represents the sample interval and was taken after the addition of the ~1cm$^3$ of Martian regolith to the WCL cells.

Previous analysis of the WCL data has employed Fourier filtering [2] and Kalman smoothing [3] techniques to reduce the noise associated with the data sets, under the assumption that the associated noise is mostly random in nature. However, through inspection of the data, we observe that for much of the data, the potential readings of the different ISEs vary simultaneously and in a similar manner, although to varying degrees. This apparently systematic variation among sensors within the same beaker, and therefore subject to the same environmental conditions, lead us to believe that these deviations could be isolated and removed from the true signal in order to reduce the uncertainty in the measured concentration of each ion and provide meaningful quantitative results from the WCL ISE analyses.

### 4.2 Number of Common Factors

The appropriate number of common factors to use in our algorithm is determined by considering the reduction in the standard error associated with the introduction of an additional common factor (strategy is described in Section 2). Fig. 6 displays the standard error for each ISE on each sol as a function of the number of common factors applied.

An overall reduction in standard error occurs for the majority of ISEs across all three sols with the addition of the first two common factors. Yet, upon the addition of the third common factor only one ISE error reduction is observed, suggesting the use of two common factors in our method.

The use of two common factors also make intuitive sense if we consider that the common variations are likely produced through two primary sources: electronic factors due to instrument malfunction, and physical factors relating to the combined effects of the physical environment inside the beaker. Therefore, we employ a two-common-factor algorithm to the WCL ISE data.

a) A reduction in variation compared to the original data.

b) Automated removal of spikes in the signal that occur simultaneously in multiple sensors.

c) Minimal deviation in mean potential from the original analysis, except for measurements in which large common variations significantly impact the mean.

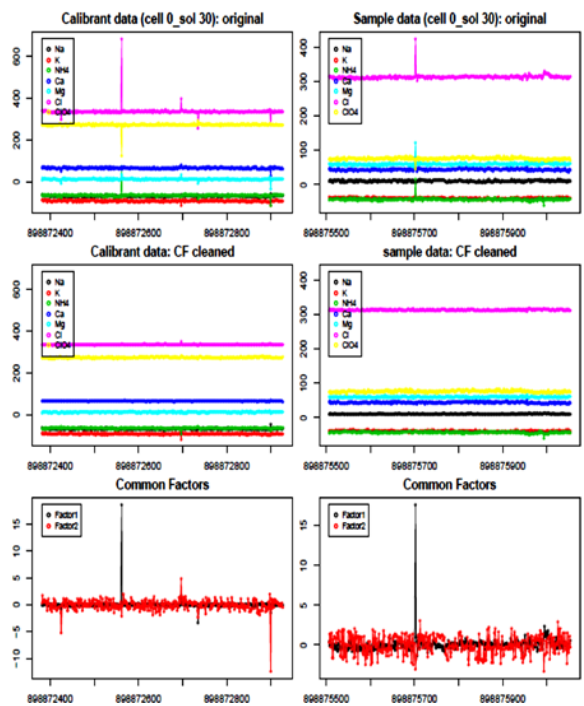

Figure 7. For Cell 0 sol 30 data (better viewed in color). Top: ISE measured potential vs Time (SCLK) for original data; middle: ISE measured potential vs Time for common-factor cleaned data; bottom: the two common factors found. From left to right: for calibrant data, sample data.

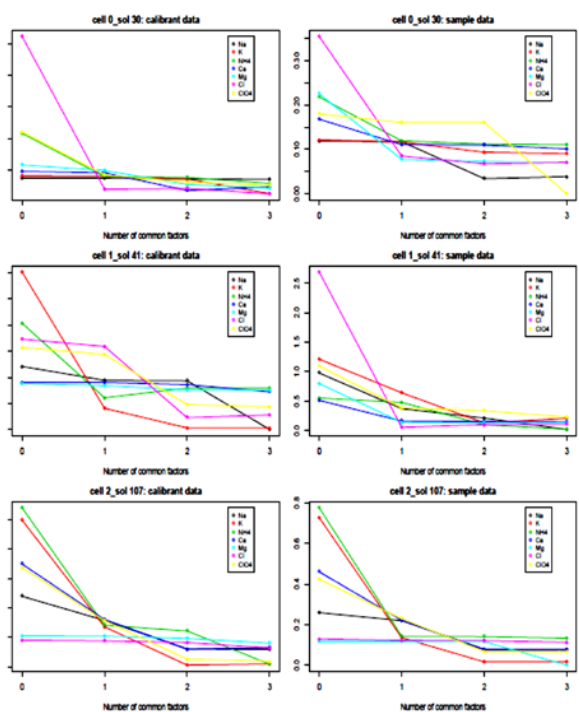

**Figure 6. Standard error vs. number of factors.**

### 4.3 Common-factor Data Cleaning

The two-common-factor algorithm was applied to the calibration and sample intervals of the WCL ISE data from cells 0, 1, and 2. The unprocessed data, the common-factor cleaned data, and the extracted common factors for each cell are displayed in Figures 7-9. The two-common-factor algorithm application to the WCL ISE measurements resulted in:

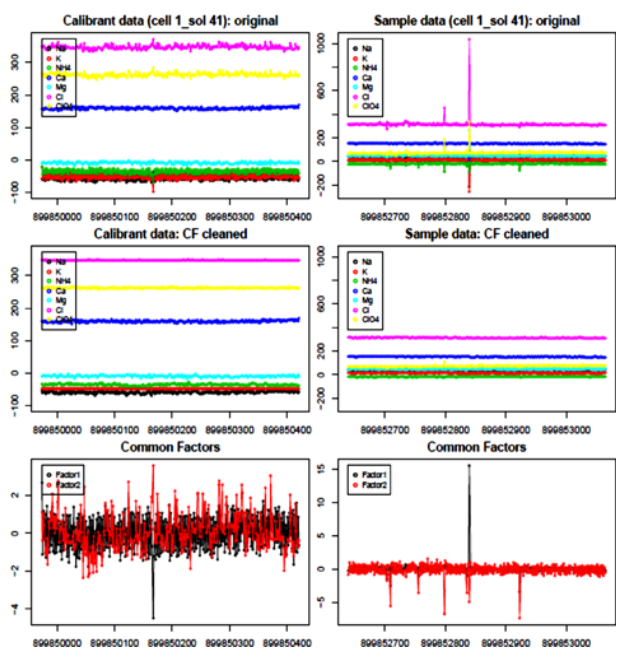

**Figure 8.** For Cell 1 sol 41 data. Same description as in Figure 7.

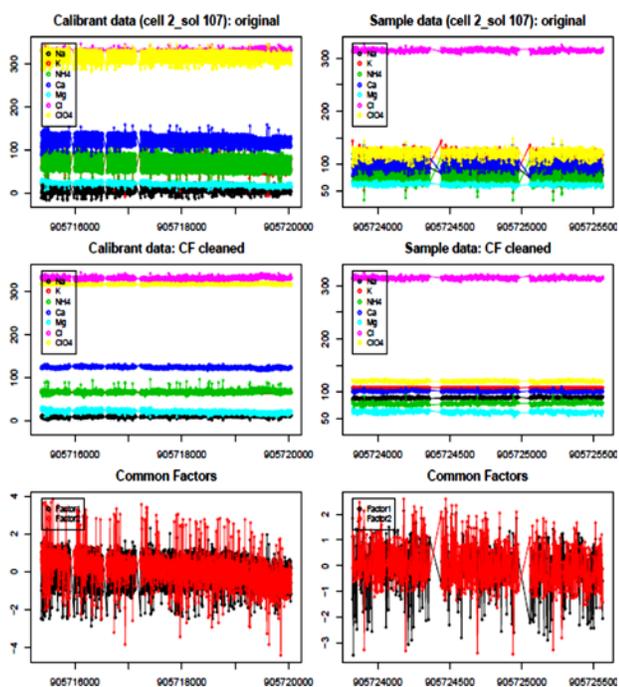

**Figure 9.** For Cell 2 sol 107 data. Same description as in Figure 7.

Using the common-factor cleaned data, the total ion concentrations and associated uncertainty are calculated (Table 1 and Figure 10) and compared with concentration estimates from previous studies using Fourier filtering by Kounaves et al [2] and Kalman smoothing by Toner et al [3]. The new ion concentrations are determined using the mean of the common-factor cleaned data and the uncertainty was generated from the standard deviations for the newly cleaned data and the error values given by the original interpretation using the standard error propagation equation

$$\sigma^2(f(x, y, \dots)) = \left(\frac{\partial f}{\partial x}\sigma_x\right)^2 + \left(\frac{\partial f}{\partial y}\sigma_y\right)^2 + \cdots$$
(8).

The highlighted values are the estimated total concentrations that are outside of the previously calculated concentration ranges. Highlights in red are ion concentrations outside Kounaves et al.'s estimated range [2], and blue highlighted values are outside Toner et al.'s estimated range [3]. The original analysis of the WCL data by Kounaves et al. used asymmetric errors to address the apparent bias in the signal noise. We report a symmetric error for our analysis, as our common-factor algorithm automatically accounts for this bias allowing the use of the standard error as an estimate of the uncertainty.

The concentration ranges estimated by our common-factor method overlap with the ranges estimated by the original Fourier filtering method for most ions in Cell 0 and Cell 1. The notable exception is the case of the $Ca^{2+}$ ISE which is reported by our common-factor method to be significantly less than originally determined for all analyses. This extreme deviation is due to the effect of the presence of $ClO_4^-$ on the $Ca^{2+}$ sensor. As described in [2], the potential used to determine the concentration of $Ca^{2+}$ is calculated based not only on the potential measured by the $Ca^{2+}$ sensor, but also the concentration of $ClO_4^-$ and the resulting changes in the reported $Ca^{2+}$ concentration is affected by altered values for both measurements. For Cell 0, all three estimates agree primarily except $Ca^{2+}$. For Cell 1, while our results are primarily in agreement with those reported by Kounaves et al., they differ dramatically from those reported by Toner et al. This is likely due to the handling of complications that arose during the initial calibration period where it is believe that the calibrant crucible intended to deliver a known concentration of ions to the leaching solution did not fully dissolve [2].

The analysis conducted in Cell 2 (Sorceress 2), shows the opposite trend, wherein our new concentration estimates vary significantly from Kounaves et al, yet agree with Toner et al. Deviation from the originally reported values in this case is not unexpected as the data returned from the Sorceress 2 analysis exhibited the largest degree of noise. That our newly reported values agree with the results published by Toner, suggests that these values are reasonable recalculations based on the denoised data set.

| Cell 0 | Concentration (M) | | | | | | |
|---|---|---|---|---|---|---|---|
| | Common Factor | | Fourier Filtering | | | Kalman Smoothing | |
| | average | ± | average | + | − | average | ± |
| Na+ | 1.50E-03 | 8.93E-05 | 1.44E-03 | 6.50E-04 | 4.80E-04 | 1.46E-03 | 3.30E-04 |
| K+ | 3.58E-04 | 2.29E-05 | 3.55E-04 | 2.90E-04 | 1.70E-04 | 3.30E-04 | 5.00E-05 |
| Ca2+ | 1.42E-04 | 1.04E-06 | 5.53E-04 | 7.50E-04 | 3.40E-04 | 1.60E-04 | 7.00E-05 |
| Mg2+ | 2.80E-03 | 9.87E-05 | 2.93E-03 | 1.90E-03 | 1.20E-03 | 2.91E-03 | 8.50E-04 |
| Cl- | 6.00E-04 | 5.30E-06 | 6.05E-04 | 1.40E-04 | 1.20E-04 | 3.90E-04 | 4.00E-05 |
| ClO4- | 2.70E-03 | 2.82E-04 | 2.64E-03 | 1.40E-03 | 9.50E-04 | 2.89E-03 | 5.40E-04 |
| NH4+ | 5.02E-05 | 2.49E-06 | 4.30E-05 | 4.20E-05 | 3.20E-05 | | |
| Cell 1 | CF | | JGR | | | Toner | |
| | average | ± | average | + | − | average | ± |
| Na+ | 1.10E-03 | 4.40E-05 | 1.10E-03 | 6.00E-04 | 3.80E-04 | 3.52E-03 | 4.50E-04 |
| K+ | 1.49E-04 | 8.23E-06 | 1.65E-04 | 2.00E-04 | 9.80E-05 | 5.00E-04 | 1.70E-04 |
| Ca2+ | 1.15E-04 | 7.30E-07 | 4.28E-04 | 7.60E-04 | 3.10E-04 | 4.50E-04 | 1.80E-04 |
| Mg2+ | 2.20E-03 | 1.16E-04 | 2.24E-03 | 2.00E-03 | 1.10E-03 | 6.22E-03 | 2.23E-03 |
| Cl- | 2.66E-04 | 1.85E-06 | 2.41E-04 | 1.30E-04 | 1.10E-04 | 7.90E-04 | 1.40E-04 |
| ClO4- | 2.10E-03 | 2.17E-04 | 2.06E-03 | 1.20E-03 | 8.60E-04 | 2.11E-03 | 5.00E-04 |
| NH4+ | 5.75E-05 | 2.66E-06 | ND | | | | |
| Cell 2 | CF | | JGR | | | Toner | |
| | average | ± | average | + | − | average | ± |
| Na+ | 1.20E-03 | 6.59E-05 | 1.44E-03 | 1.00E-03 | 6.10E-04 | 9.90E-04 | 2.80E-04 |
| K+ | 2.07E-04 | 4.42E-06 | 3.87E-04 | 3.20E-04 | 1.70E-04 | 1.70E-04 | 3.00E-05 |
| Ca2+ | 1.51E-04 | 1.74E-07 | 6.03E-04 | 7.90E-04 | 3.40E-04 | 9.00E-05 | 4.00E-05 |
| Mg2+ | 1.60E-03 | 3.87E-05 | 3.70E-03 | 3.00E-03 | 1.70E-03 | 1.31E-03 | 4.20E-04 |
| Cl- | 4.64E-04 | 2.27E-06 | 4.63E-04 | 2.10E-04 | 1.10E-04 | 2.40E-04 | 3.00E-05 |
| ClO4- | 2.50E-03 | 1.16E-04 | 2.15E-03 | 2.20E-03 | 8.10E-04 | 2.72E-03 | 5.70E-04 |
| NH4+ | 2.99E-05 | 6.19E-07 | 2.60E-05 | 6.00E-05 | 3.20E-05 | | |

**Table 1: Estimated total ion concentrations with associated errors calculated with our common-factor method, with Fourier filtering (from the original analysis), and using Kalman smoothing (as in Toner et al). Highlighted cells fall outside of the error range reported in the initial analysis.**

## 5. CONCLUSION AND FUTURE DEVELOPMENT

In this paper, we present a new common-factor method for reducing unwanted variations from common interferences in data signals. The method is easy to use, intuitive and effective as a more unified approach for cleaning data. Our method eliminates the need for special handling of data in complicated scenarios where the origin of the noise is difficult to understand.

To date this common-factor algorithm has been successfully applied to the WCL experiments, as demonstrated, and would likely prove successful in other cases of data interpretation where the results are linked together in a sensor array and subject to extensive but unknown systematic noise. Sensor arrays are commonly employed in the field of environmental monitoring, wherein several different measurements are obtained simultaneously and the combination of data is used to obtain otherwise inaccessible information about the system.

When these sensor arrays are employed in extreme and remote environments, the data obtained may exhibit extensive noise that, while unknown in source, affects all individual sensors to varying degrees. This new common-factor method can aid in reducing this systematic noise without a definitive understanding of the source and without degrading the physical meaning of the signal.

This work is important to scientific discoveries because of the following.

- A method of removing common systematic error of unknown source can be implemented in data analysis for similar missions. This is paramount for analyses performed in extreme and extraterrestrial environments as unanticipated and unknown factors affecting data measurements are common.
- The ability to analyze the data output from a sensor array for common variations that are independent of the chemistry provides the opportunity for gathering future data in complex samples where many unknown contributions to the signal exist.
- A cleaner data set for the WCL analysis provides reduced uncertainty in the soluble chemistry of the Martian regolith. This will allow for more accurate geochemical models to be constructed and lead to a greater degree of certainty in the interpretations of the data.

## 6. ACKNOWLEDGMENTS

This work was supported by NASA under Grant NNX13AJ69G.

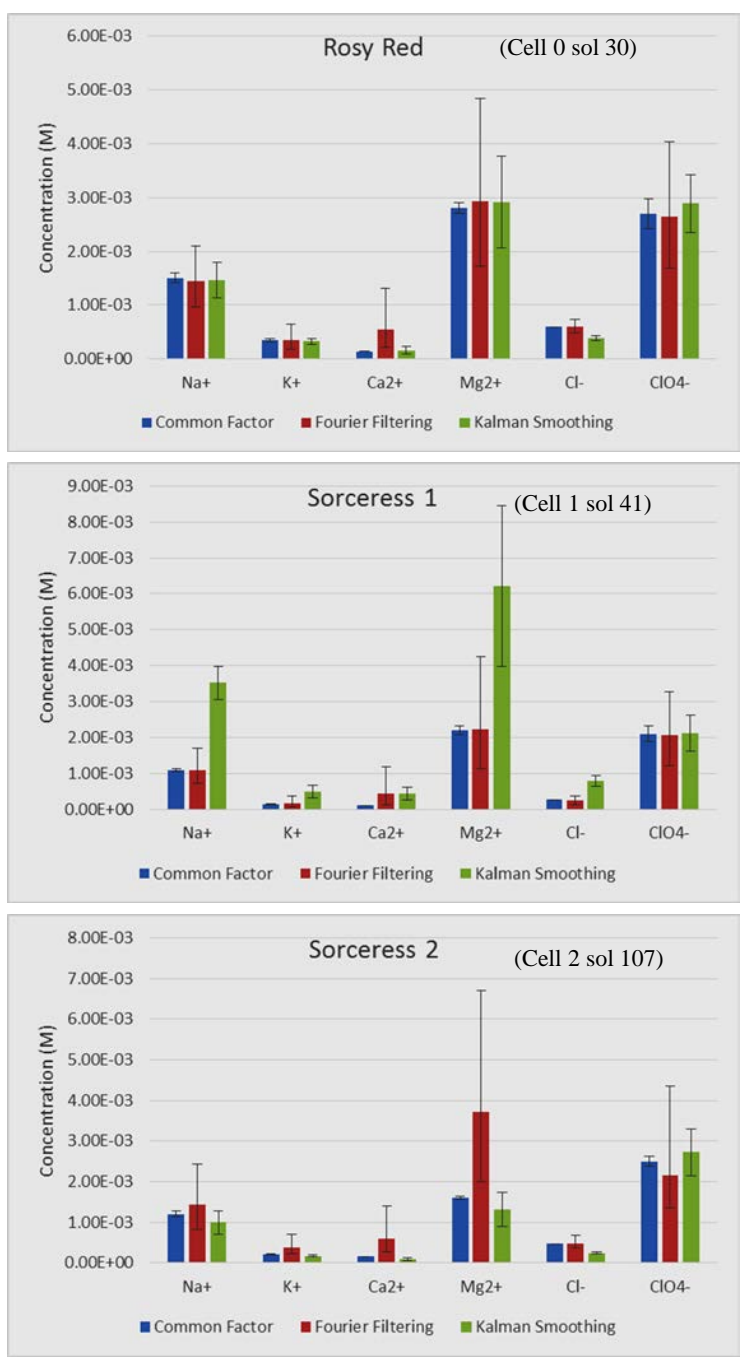

**Figure 10. Estimated total ion concentrations and their associated error bars.**